\PassOptionsToPackage{numbers,sort&compress}{natbib}
\documentclass[sigconf,nonacm]{acmart}

\usepackage[utf8]{inputenc}
\usepackage[T1]{fontenc}
\usepackage{microtype}
\usepackage{booktabs}
\usepackage{tabularx}
\usepackage{array}
\usepackage{amsmath,amssymb}
\usepackage{graphicx}
\usepackage{xurl}
\usepackage{xcolor}

\setcopyright{none}
\hypersetup{
  hidelinks,
  pdftitle={When Are Aggregate Agent Traces Diagnosable? Traffic-Governed Interpretation and Calibrated Abstention},
  pdfauthor={Peiying Zhu; Sidi Chang}
}

\providecommand{\tightlist}{\setlength{\itemsep}{0pt}\setlength{\parskip}{0pt}}
\begin{document}
\acmshorttitle{When Are Aggregate Agent Traces Diagnosable?}
\acmshortauthors{Zhu \& Chang}
\pagestyle{acmstyle}
\thispagestyle{firstpage}

\twocolumn[
\begin{@twocolumnfalse}
\begin{center}
{\LARGE\bfseries When Are Aggregate Agent Traces Diagnosable? Traffic-Governed Interpretation and Calibrated Abstention\par}
\vspace{1.0em}

\begin{tabular}{@{}c@{\hspace{2em}}c@{}}
{\large Peiying Zhu\textsuperscript{*}} & {\large Sidi Chang\textsuperscript{*\,$\dagger$}} \tabularnewline
{\small peiying@blossomai.co} & {\small schang@blossomai.co} \tabularnewline
{\small Blossom AI} & {\small Blossom AI} \tabularnewline
{\small San Francisco, CA, USA} & {\small San Francisco, CA, USA}
\end{tabular}
\vspace{1.2em}
\end{center}

\noindent\parbox{\textwidth}{
\small
\noindent\textbf{Abstract}\\[3pt]
Runtime traces are often treated as transparent evidence about an agent, but a closed-loop policy determines which states are visited and therefore which failures can become visible. We study a simulated hotel-pricing agent whose policy maps time, inventory, and market state to discrete price actions under varying demand regimes. A fault may leave no aggregate trace when the policy rarely visits its affected cells. We treat entry into aggregate-only fault interpretation as a diagnosability decision that precedes scoring or localization. A reference-map gate requires repeated clean-policy support; a matched runtime gate then requires joint support in clean and current streams. Stable signal analysis occurs only after both gates pass. We calibrate false admission on a disjoint clean stream at the physical-component level and model detection by affected clean traffic rather than nominal cell coverage.

In a frozen one-shot heldout, 55/72 (76.4\%) regime-component units were reference-admitted, representing 20 physical components; 54/55 then passed matched runtime admission, and the rejected unit abstained. Stable false admission was 0/20, with one-sided exact 95\% upper bound 0.1391, meeting the frozen 0.20 criterion. Across 540 repeated unit-arm rows nested in those 20 clusters, affected clean traffic reduced negative log likelihood by 29.3\% relative to cell coverage, a gain of 0.1264 nats per row (cluster-bootstrap 95\% interval {[}0.0593, 0.1918{]}). Adding mask family and its interaction improved log loss by only 0.0015 nats per row (one-sided upper bound 0.0066), below the frozen 0.01 practical-sufficiency margin. A development audit also found that exact minimum hitting set and greedy selection chose identical supports in 12/12 scenarios because singleton evidence had already resolved the conflicts. The result is a bounded rule for interpreting aggregate agent behavior: first establish exposure, then score change, and abstain when the trace cannot support the claim.

}

\vspace{1.5em}
\end{@twocolumnfalse}
]

\renewcommand{\thefootnote}{\fnsymbol{footnote}}
\footnotetext[1]{Both authors contributed equally to this research.}
\footnotetext[2]{Corresponding author.}
\renewcommand{\thefootnote}{\arabic{footnote}}

\section{Introduction}\label{introduction}

Agent evaluation is moving from final-task scores toward trajectories: tool calls, intermediate states, recovery behavior, and long-horizon traces \cite{manualRef18,manualRef19,manualRef20}. These records are richer than an outcome, but they are not neutral windows into the system. Closed-loop agents act on state, and those actions change which states will be observed next. A component can therefore fail without producing a visible aggregate change when the policy rarely visits the affected region. Demand shift can also change visitation and action summaries without any component fault. No diagnosis algorithm can recover information absent from its measurements \cite{manualRef03}; trace interpretation that ignores exposure can turn missing evidence into reassurance or ordinary drift into a fault.

The system is a simulated hotel-pricing agent whose policy maps time, inventory, and market state to discrete price actions under varying demand regimes. Its target policy is partitioned into 24 behavioral components, and the interpreter receives typed aggregate trajectories rather than the planted fault identity or every internal decision. Earlier versions treated exact, risk-calibrated minimum hitting set (MHS) as the main contribution. Conflict-directed diagnosis maps conflicts to hitting-set candidates \cite{manualRef01,manualRef02}, but our audit changed that conclusion: exact MHS and weighted greedy selected the same support in all 12 development scenarios. Exact-scope predicates and one-component probes had already introduced planted-component singletons, leaving no residual ambiguity after propagation.

This negative result redirects the question upstream. If a trace compiler already reveals the answer, a stronger optimizer cannot demonstrate a localization advantage. If policy traffic never exposes the affected region, neither exact nor approximate optimization can recover it. The consequential interpretive decision is whether the current trace contains evidence about the claimed component at all.

Our workflow uses two gates. The reference-map gate asks whether a component is repeatedly visited in a clean stream. The runtime gate asks whether clean and current streams are jointly supported in matched partitions. Only then is stable signal evaluated. Clean-versus-clean false admission is calibrated separately, and the physical component, not each repeated regime arm or simulator trajectory, is the safety sampling unit.

We also measure fault severity in the coordinates that generate the traces. A mask may change many unvisited cells or a few high-traffic cells, so cell count need not predict signal strength. We define affected clean traffic as the fraction of clean visits falling in changed cells, compare it with affected cell count, and test whether traffic remains practically sufficient across uniform and flow-weighted masks.

A frozen shift detector separately triggers map recomputation rather than declaring a fault or an abstention. This keeps distribution change, support loss, and stable fault signal as distinct events.

The paper makes four bounded contributions to interpreting agent behavior:

\begin{enumerate}
\def\labelenumi{\arabic{enumi}.}
\tightlist
\item
  We operationalize trace diagnosability with reference and matched-runtime support gates that return explicit abstentions.
\item
  We define affected clean traffic as a behavior-grounded exposure variable and test it against nominal edit size.
\item
  We pre-register independent false-admission calibration and family-robustness criteria at the physical-component level.
\item
  We report a negative optimizer result that exposes a verifier artifact: exact MHS showed no advantage over propagation-aware greedy selection in this benchmark.
\end{enumerate}

The heldout confirms gates, abstention, and traffic-indexed signal behavior. It does not confirm localization accuracy.

\section{Related work and conceptual position}\label{related-work-and-conceptual-position}

Interactive-agent benchmarks increasingly retain trajectories and environment state rather than only final answers \cite{manualRef18,manualRef19,manualRef20}. Such benchmarks establish tasks and observable records; they do not automatically establish that a particular trace statistic identifies a component-level failure. Our contribution is complementary: it asks when an aggregate trajectory is eligible to support a fault interpretation in the first place.

Conflict-directed model-based diagnosis derives candidates as hitting sets of conflicts \cite{manualRef01}, and exact methods improve search efficiency \cite{manualRef02}. These solvers are conditional on the compiled conflict family; they do not establish whether upstream observations distinguish the represented physical faults. Diagnosability work instead asks whether faults can be distinguished from available measurements \cite{manualRef03,manualRef04,manualRef05}. Our gates are a finite-sample operationalization for aggregate closed-loop traces, not a general diagnosability theorem.

Selective prediction trades coverage against error by rejecting unsupported inputs \cite{manualRef06,manualRef07,manualRef08,manualRef16}. Overlap analysis and offline reinforcement learning similarly avoid inference where state-action coverage is weak \cite{manualRef09,manualRef10,manualRef11,manualRef12}. We apply this support-first principle before trace interpretation, with fault visibility rather than treatment effect or policy value as the estimand. Dataset-shift and conformal methods motivate a separate drift-to-refresh path \cite{manualRef13,manualRef14,manualRef15}. Recent work interprets runtime behavior through pointwise temporal-logic diagnostics \cite{manualRef21} and trajectory-level competence and visualization tools for LLM agents \cite{manualRef22}; complementary theory establishes limits on predicting agents from behavior alone \cite{manualRef23}. Our narrower question is whether an aggregate trace is eligible to support a component-level fault interpretation. Our contribution is the empirical synthesis in a pre-localization pipeline: support gates, component-level false-admission calibration, and traffic-indexed signal analysis. We do not claim to originate diagnosability, abstention, overlap analysis, shift detection, or behavioral diagnostics.

\section{Problem setting}\label{problem-setting}

\subsection{Components, regimes, and aggregate traces}\label{components-regimes-and-aggregate-traces}

Let \texttt{C} be the set of 24 disjoint target-policy components. Each component is indexed by a time quarter, inventory half, and market third. The simulator is evaluated under three separately fitted demand regimes, \texttt{lambda0\ in\ \{5,\ 7,\ 9\}}. A regime-component pair is a unit, giving 72 units before admission.

For each unit and evaluation partition, the system records typed aggregate traces. The signal calculation uses two existing summaries:

\begin{itemize}
\tightlist
\item
  \texttt{region\_d1}, a regional distribution distance; and
\item
  \texttt{mean\_action\_gap}, the mean reference-current action difference.
\end{itemize}

The diagnostic does not observe the planted fault identity when deciding admission or signal stability.

\subsection{Fault masks and directions}\label{fault-masks-and-directions}

A formal fault changes a deterministic subset of direction-changeable target-field cells by one action bucket. Each selected subset is tested in outward and inward directions. Two nested mask families are used:

\begin{itemize}
\tightlist
\item
  \texttt{uniform}, a seeded random ordering of changeable cells; and
\item
  \texttt{flow\_weighted}, a seeded weighted-without-replacement ordering in which frequently visited clean cells tend to enter earlier.
\end{itemize}

The two families can have the same selected cell fraction but very different affected traffic.

\subsection{Affected clean traffic}\label{affected-clean-traffic}

For component cell \(c\), let \(o_c\) be its clean reference visit count. For a directional fault selecting cells \(S_d\), define

\[\tau_d = \frac{\sum_{c \in S_d} o_c}{\sum_{c \in \text{component}} o_c}.\]

The primary exposure is

\[\tau = (\tau_{\mathrm{outward}} + \tau_{\mathrm{inward}})/2.\]

For the matched cell-count model, directional coverage is the selected fraction of direction-changeable cells, and the bi-directional covariate is the arithmetic mean

\[f_{\mathrm{cell}} = (f_{\mathrm{outward}} + f_{\mathrm{inward}})/2.\]

If component occupancy is zero, \texttt{tau} is zero. Zero-exposure cases are reported separately and are not shifted by an arbitrary constant before logging.

\subsection{Diagnosability as a state, not a hidden assumption}\label{diagnosability-as-a-state-not-a-hidden-assumption}

For a unit \texttt{u}, a reference stream \texttt{R}, and a current stream \texttt{Q}, the diagnostic state is:

\[
\begin{gathered}
\texttt{REFERENCE\_ABSTAIN} \rightarrow \texttt{RUNTIME\_ABSTAIN} \\
\rightarrow \texttt{SIGNAL\_ELIGIBLE} \\
\rightarrow \{\texttt{DETECTED},\texttt{NOT\_DETECTED}\}.
\end{gathered}
\]

The arrows denote tests, not a temporal guarantee that every unit progresses. A unit that fails the first gate never reaches the second. A unit that fails the second never reaches signal interpretation. This ordering distinguishes ``no usable comparison'' from ``usable comparison with no stable signal.''

\section{Why MHS is downstream}\label{why-mhs-is-downstream}

Given conflicts and component costs, exact MHS finds the minimum-cost set intersecting every conflict. Its guarantee remains conditional on the conflict family and does not establish physical-fault identification. In 12 audited development scenarios, exact MHS and propagation-aware greedy selection chose identical supports and achieved identical exact recovery (9/12). The instances were not devoid of combinatorial structure: non-singleton conflicts comprised 55.5\% of the conflict collection, and conflicts overlapped in all 12 scenarios. A follow-up audit found planted-component singleton conflicts in 9/9 exact-anchor and 9/9 hard-probe cases, with 0/9 nonempty residual conflict families after propagation.

The mechanism is upstream. Classical GDE selects a next measurement to discriminate among diagnoses by expected entropy reduction \cite{manualRef17}. The hard probe here instead changed exactly one component and paired that intervention with a predicate scoped to the same component, converting the local response into a planted-component singleton. Propagation therefore forced the choice before exact and greedy solvers could differ. This does not show that MHS is wrong; it shows that the evidence construction does not test optimizer advantage. We retain MHS only as an optional downstream risk-cost optimizer and make no formal localization claim.

\section{Diagnosability method}\label{diagnosability-method}

\subsection{Reference-map gate}\label{reference-map-gate}

A component has partition support when its existing support count is at least 12. A regime-component unit enters the formal reference map when at least 14 of 15 clean reference partitions have support. All 72 units are screened; none is excluded by identity or development eligibility.

The admitted proportion over 72 units is the primary coverage description. The number of distinct physical components represented is also reported. A low admission rate is a limitation of the observable system, not a reason to change the gate.

\subsection{Runtime two-stream gate}\label{runtime-two-stream-gate}

A runtime partition is supported only when reference and current support are both at least 12 in that same partition. Runtime admission requires at least 14 such joint partitions out of 15. For a bi-directional fault observation, both directions must pass.

This intersection matters because separate aggregate counts can hide mismatched evidence: 14 well-supported reference partitions and 14 well-supported current partitions do not guarantee 14 usable comparisons if they occur in different partitions.

\subsection{Stable aggregate signal}\label{stable-aggregate-signal}

For each matched reference-current partition pair, define

\[z = \max\!\left(\frac{\texttt{region\_d1}}{0.20},
\frac{|\texttt{mean\_action\_gap}|}{0.35}\right).\]

A partition signals when \texttt{z\ \textgreater{}=\ 1.50}. After the unit-level runtime gate passes, direction stability is counted over all 15 matched pairs rather than over a support-signal intersection; a direction is stable at 14/15 signaling partitions, and a unit-arm is bi-directionally stable only when both directions are stable. Signal values for an arm that fails either support gate remain sealed validation fields and cannot enter a power endpoint. Threshold 1.50 was selected before formal execution from the fixed grid \texttt{\{1.00,\ 1.25,\ 1.50,\ 1.75,\ 2.00\}}. It is not tuned on the formal null.

\subsection{Three separate failure quantities}\label{three-separate-failure-quantities}

We separate:

\begin{enumerate}
\def\labelenumi{\arabic{enumi}.}
\tightlist
\item
  \textbf{reference-map invalidation:} a development-admitted reference unit fails the initial formal reference-only rule;
\item
  \textbf{runtime two-stream rejection:} the formal reference passes but the matched current-reference comparison fails joint support; and
\item
  \textbf{stable false admission:} a runtime-admitted clean-versus-clean unit produces a stable signal.
\end{enumerate}

Combining them would obscure whether the problem is a changed reference map, insufficient concurrent evidence, or a noisy signal. Each is reported separately at unit and physical-component-cluster levels. Reference invalidation and runtime rejection are descriptive coverage endpoints with no frozen population-rate threshold; each affected unit still abstains. The only \(\le 0.20\) population claim is stable false admission.

\subsection{Drift and refresh}\label{drift-and-refresh}

A frozen split-conformal detector summarizes each clean-null partition and raises an alarm above a precommitted threshold. Alarms request atomic map recomputation; they do not declare a fault or an abstention. Passing units return to \texttt{MAP\_VALID}, while units failing the refreshed reference rule enter \texttt{REFERENCE\_ABSTAIN}. The formal endpoint is only a state-machine smoke test because every refresh reuses the same frozen reference buffer; it does not show that future data can repair a stale map.

\section{Frozen confirmatory design}\label{frozen-confirmatory-design}

\subsection{Seed separation}\label{seed-separation}

Formal seeds \texttt{200000..202399} form 15 clean reference partitions of 160 episodes. Disjoint seeds \texttt{202400..204799} form 15 current/fault partitions. Seeds \texttt{204800..204999} remain unused. Common random numbers pair conditions within the experiment, while inference clusters repeated observations by physical component.

An episode is one simulator trajectory used to form a partition-level aggregate; it is not an independent statistical unit. The 3,456,000 scheduled fault-stream episodes describe Monte Carlo workload, not an effective sample size. The design begins with 24 physical components repeated across three regimes, producing 72 coverage units. After admission, safety and bootstrap inference operate on 20 represented physical-component clusters. The primary structural fit contains 540 unit-arm rows nested within those clusters.

\subsection{Formal mask construction}\label{formal-mask-construction}

For each regime, component, direction, and family, a SHA-256-derived seed produces a deterministic cell ordering under master seed \texttt{2026082370}. Nested prefixes target traffic near 0.15, 0.30, 0.50, 0.75, and 0.95; achieved traffic enters analysis. The clean stream determines occupancy and masks before fault outcomes are opened, and selected-cell, traffic, cell-fraction, and policy hashes are sealed.

\subsection{Formal endpoints}\label{formal-endpoints}

Endpoints cover admission, component-level reference invalidation and runtime rejection, component-level stable false admission, stable-signal power versus achieved traffic, traffic-versus-cell log-loss, practical family sufficiency, aggregate monotonicity, and state-machine execution. MHS output and planted localization accuracy are excluded.

\subsection{Statistical analysis}\label{statistical-analysis}

Reference invalidation and runtime rejection use component-level two-sided exact 95\% Clopper-Pearson intervals as descriptive coverage endpoints; they have no population-rate pass threshold. Stable false admission alone uses the component-level one-sided exact 95\% upper bound and is labeled \texttt{SAFETY\ CONFIRMED} only when that bound is at most 0.20. Regime-unit rates are secondary because regimes reuse the same physical components.

The safety rule has coarse resolution: at 20 admitted components, \texttt{0/20} gives upper bound 0.1391 and passes, while \texttt{1/20} gives 0.2161 and fails.

Power rates use Wilson 95\% intervals. The primary model fits stable signal against \texttt{log(tau)} on positive-exposure units that pass both support gates. A matched cell model uses \texttt{log(cell\_fraction)}. Traffic is superior only if the component-bootstrap two-sided 95\% lower bound on per-unit-arm-row negative-log-likelihood improvement, \texttt{NLL(cell)\ -\ NLL(traffic)}, is greater than zero.

The family model adds mask family and its interaction with \texttt{log(tau)}. Traffic is practically sufficient across the two frozen families only if the component-bootstrap one-sided 95\% upper bound on augmented-model log-loss improvement is below 0.01 nats per unit-arm row. A nonsignificant family coefficient is not an equivalence result.

The bootstrap samples physical components, not unit-arm rows. A fixed generator produces 2,000 draws; finite singular and one-class replicates are retained, and no replicate is redrawn. Empty or nonfinite analyses return \texttt{NOT\ ESTIMABLE} rather than a favorable verdict.

The exact-binomial safety statement is conditional on the frozen working assumption that distinct physical-component indicators are exchangeable independent Bernoulli trials given the policies and shared seed schedule. Common random numbers can induce residual cross-component dependence, so this is not an unconditional new-population guarantee.

\subsection{Development separation}\label{development-separation}

Development data motivated traffic as the primary axis but did not establish family sufficiency. They were not pooled with the formal heldout. A pre-run projection from 20 to 15 partitions and the withdrawn cell-coverage heuristic were development-only; neither is a formal endpoint.

\section{Formal results}\label{formal-results}

\subsection{Sampling frame, admission, and abstention}\label{sampling-frame-admission-and-abstention}

Of 72 regime-component units, 55/72 passed formal reference admission, representing 20/24 distinct components. Relative to the development reference map, invalidation was 0/20 at component grain and 0/55 at regime-unit grain; the component-level descriptive two-sided exact 95\% interval was {[}0.0\%, 16.8\%{]}. On the clean current stream, 1/20 component clusters had a runtime two-stream rejection, with descriptive interval {[}0.1\%, 24.9\%{]}. These two rates quantify coverage loss and have no population pass threshold; every affected unit abstained.

Among the 20 runtime-admitted components, 0/20 produced a stable clean-versus-clean signal. The one-sided exact 95\% upper bound was 0.1391, satisfying the frozen \texttt{\textless{}=0.20} safety rule. The corresponding regime-unit result was 0/54 and remains descriptive.

\begin{table*}[t]
\caption{Formal admission, abstention, and false-admission endpoints.}
\centering
\small
\begin{tabularx}{\textwidth}{@{}>{\raggedright\arraybackslash}p{0.13\textwidth}>{\raggedright\arraybackslash}p{0.16\textwidth}>{\raggedleft\arraybackslash}p{0.14\textwidth}>{\raggedright\arraybackslash}X>{\raggedright\arraybackslash}X@{}}
\toprule
Grain & Endpoint & Result & Interval / bound & Interpretation \\
\midrule

regime-component & reference admission & 55/72 (76.4\%) & descriptive & addressable units \\
physical component & reference representation & 20/24 (83.3\%) & descriptive & addressable components \\
physical component & reference invalidation & 0/20 & 95\% CI {[}0.0\%, 16.8\%{]} & descriptive; abstain \\
physical component & runtime rejection & 1/20 & 95\% CI {[}0.1\%, 24.9\%{]} & descriptive; abstain \\
physical component & stable false admission & 0/20 & one-sided 95\% upper 0.1391 & \textbf{SAFETY CONFIRMED} \\
\bottomrule
\end{tabularx}
\end{table*}

\subsection{Traffic-indexed power}\label{traffic-indexed-power}

\begin{figure*}[t]
\centering
\includegraphics[width=\textwidth]{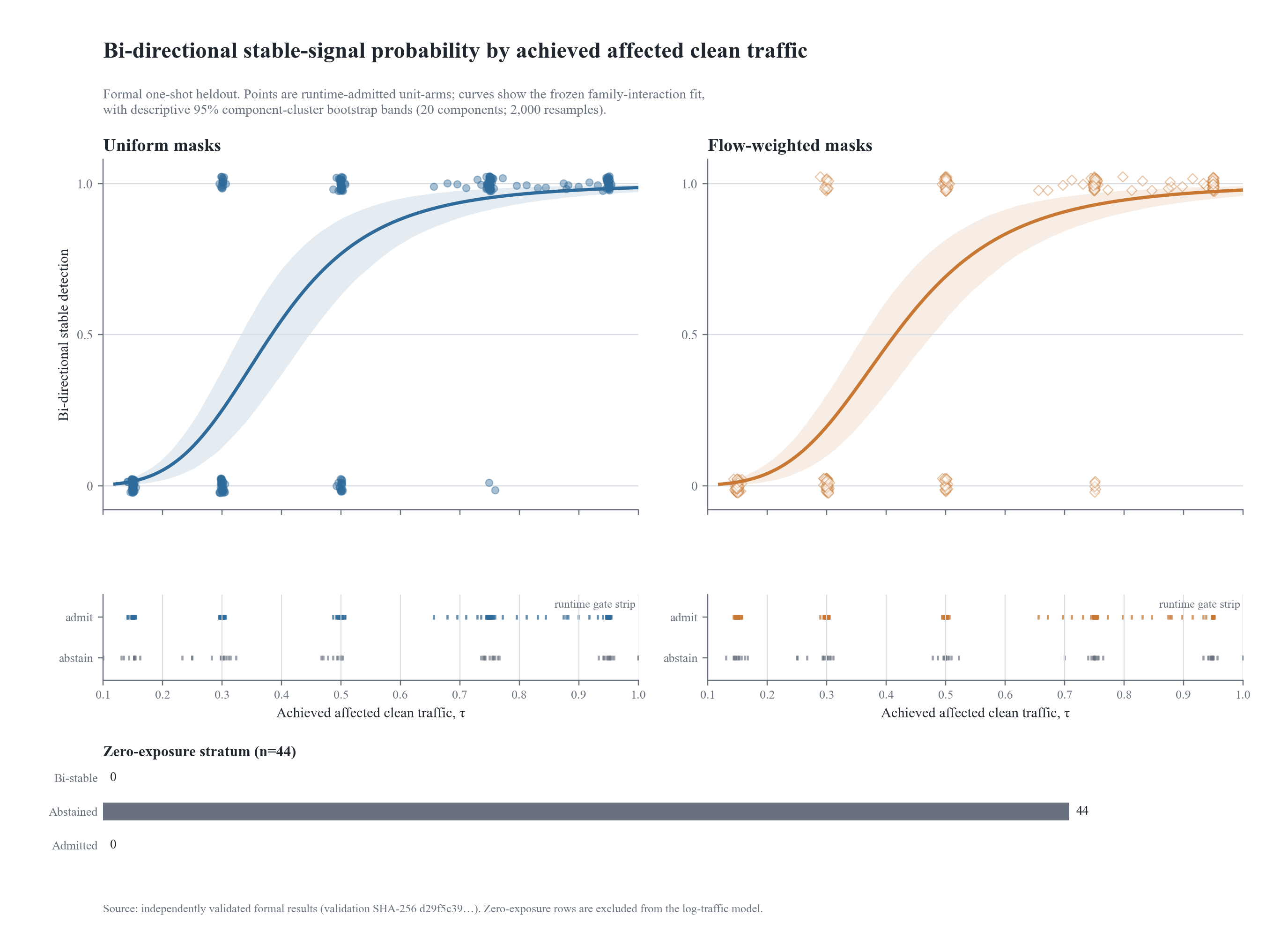}
\caption{Bi-directional stable-signal probability versus achieved affected clean traffic. Points and pre-registered family-interaction curves are shown separately for uniform and flow-weighted masks. Shaded bands are descriptive 95\% component-cluster bootstrap intervals from 20 represented components and 2,000 resamples. Runtime admission is shown below each curve; all 44 zero-exposure observations appear separately.}
\end{figure*}

\begin{table*}[t]
\caption{Conditional and operational bi-directional stable detections.}
\centering
\small
\begin{tabularx}{\textwidth}{@{}>{\raggedright\arraybackslash}p{0.22\textwidth}ccccc@{}}
\toprule
Family & Traffic target 0.15 & 0.30 & 0.50 & 0.75 & 0.95 \\
\midrule

uniform, conditional & 0/54 & 17/54 & 37/54 & 52/54 & 54/54 \\
uniform, operational & 0/55 & 17/55 & 37/55 & 52/55 & 54/55 \\
flow-weighted, conditional & 0/54 & 13/54 & 35/54 & 49/54 & 54/54 \\
flow-weighted, operational & 0/55 & 13/55 & 35/55 & 49/55 & 54/55 \\
\bottomrule
\end{tabularx}
\end{table*}

The conditional denominator is the 54 runtime-admitted units; the operational denominator is the 55 reference-admitted units and counts the one runtime abstention as no actionable diagnosis. Both families were saturated at 0/54 and 54/54 at the endpoints; only the middle-target family gaps (+4, +2, +3, uniform minus flow-weighted) are empirically informative. Monotonicity passed but is a weak shape check, not evidence of family equivalence. All 44 zero-exposure scheduled observations abstained and none was stable.

\subsection{Traffic versus cell coverage and family sufficiency}\label{traffic-versus-cell-coverage-and-family-sufficiency}

Across 540 repeated unit-arm rows nested in 20 represented physical-component clusters, the traffic model achieved negative log likelihood 164.9866 (0.3055 per row), compared with 233.2453 (0.4319 per row) for the cell-count model. The difference was 0.1264 nats per unit-arm row, a 29.3\% reduction relative to the cell-model negative log likelihood, with component-bootstrap 95\% interval {[}0.0593, 0.1918{]}. The lower bound was above zero, so traffic was confirmed as the better primary axis.

The family-plus-interaction model improved log loss by 0.0015 nats per unit-arm row. Its one-sided component-bootstrap 95\% upper bound was 0.0066, which was below the frozen 0.01 margin. Traffic therefore met the practical family-sufficiency criterion. The traffic-plus-family coefficient was -0.3618 with component-bootstrap interval {[}-0.8617, 0.0367{]} and secondary likelihood-ratio p-value 0.2019.

The largest observed family gaps occurred in the \texttt{{[}0.30, 0.45)} and \texttt{{[}0.75, 0.90)} traffic bins. They are small in aggregate log-loss terms (0.0015 nats per row) but are not excluded on the odds scale, consistent with the coefficient interval above. The 0.0034-nat margin is narrow, and the percentile cluster-bootstrap uses only 20 represented components. The sufficiency label is therefore limited to the two frozen mask families and the precommitted 0.01-nat criterion.

\begin{table*}[t]
\caption{Pre-registered structural tests.}
\centering
\small
\begin{tabularx}{\textwidth}{@{}>{\raggedright\arraybackslash}p{0.21\textwidth}>{\raggedleft\arraybackslash}p{0.12\textwidth}>{\raggedright\arraybackslash}X>{\raggedright\arraybackslash}p{0.14\textwidth}>{\raggedright\arraybackslash}p{0.10\textwidth}@{}}
\toprule
Test & Estimate & Cluster interval / bound & Criterion & Result \\
\midrule

cell minus traffic NLL, nats/row & 0.1264 & 95\% {[}0.0593, 0.1918{]} & lower \textgreater{} 0 & \textbf{PASS} \\
family+interaction gain, nats/row & 0.0015 & one-sided upper 0.0066 & upper \textless{} 0.01 & \textbf{PASS} \\
\bottomrule
\end{tabularx}
\end{table*}

An operational sensitivity counting runtime abstention as zero gave the same decisions: traffic-over-cell improvement 0.1164 {[}0.0512, 0.1835{]} and family-plus-interaction gain 0.0014 with upper bound 0.0060.

\subsection{Drift, refresh, and decision summary}\label{drift-refresh-and-decision-summary}

The detector was quiet in its matched \texttt{lambda0=7} regime (0/15 alarms) and active under demand shifts at \texttt{lambda0=5} (15/15) and \texttt{lambda0=9} (14/15). These are descriptive state-machine inputs, not a 29/45 false-alarm estimate. Five refresh operations reproduced the frozen map over 120 evaluations; because they reuse the same buffer, this does not show that refresh can repair a stale map.

\begin{table*}[t]
\caption{Frozen decision summary.}
\centering
\small
\begin{tabularx}{\textwidth}{@{}>{\raggedright\arraybackslash}p{0.11\textwidth}>{\raggedright\arraybackslash}p{0.22\textwidth}>{\raggedright\arraybackslash}X>{\raggedright\arraybackslash}p{0.13\textwidth}@{}}
\toprule
Layer & Item & Formal result & Label \\
\midrule

execution & frozen protocol and validator & 56 checks; 1,440 cases; 21,600 partition rows & EXECUTED \\
safety & stable false-admission upper \textless=0.20 & 0/20; upper 0.1391 & CONFIRMED \\
coverage & reference invalidation / runtime rejection & 0/20 / 1/20 & DESCRIPTIVE \\
structure & traffic over cell coverage & 0.1264 {[}0.0593, 0.1918{]} & PASS \\
structure & family practical sufficiency & 0.0015; upper 0.0066 & PASS \\
structure & aggregate monotonicity (weak shape check) & nondecreasing; endpoints saturated & PASS \\
\bottomrule
\end{tabularx}
\end{table*}

There is no combined scientific verdict. Execution, operational safety, descriptive coverage, and the three structural hypotheses retain separate labels and interpretations.

\section{Discussion}\label{discussion}

\subsection{Interpretation is fixed by the endpoint pattern}\label{interpretation-is-fixed-by-the-endpoint-pattern}

The endpoint pattern fixes a bounded interpretation: stable false admission supports only the component-cluster 0.20 upper-bound statement; reference invalidation and runtime rejection remain descriptive coverage losses with abstention; and a single traffic curve is practically sufficient only within admitted units and the two frozen mask families. Nondecreasing counts are a weak shape check, not family equivalence.

\subsection{What the work changes conceptually}\label{what-the-work-changes-conceptually}

Optimization and observability are distinct: exact MHS may be optimal for a compiled objective without identifying the physical fault, while a simple method can look perfect when singleton evidence already reveals the answer. Neither shows whether a trace reveals failure under a different traffic regime.

The two-gate workflow makes that dependency explicit. Reference admission asks what the policy normally exposes; runtime admission asks whether a matched comparison is possible; signal admission asks whether change is stable; and localization comes last. This is an application-specific synthesis of diagnosability analysis \cite{manualRef03,manualRef04}, local support characterization \cite{manualRef09}, and selective abstention \cite{manualRef06,manualRef07,manualRef16}.

Traffic exposure provides a mechanism for why maps move with environment. Higher demand consumes inventory more quickly, transferring visitation from high-inventory to low-inventory regions later in the episode. Pooled component ranks can therefore fall even when direction-specific movement is predictable. A map may rotate along a known inventory axis rather than become arbitrary. The operational consequence is to refresh the map when drift is detected, not to permanently reject the diagnostic.

\subsection{Why abstention is part of the result}\label{why-abstention-is-part-of-the-result}

Admission below 100\% defines the supported portion of the component universe, not merely lost sample size. Reporting conditional power alongside operational power, with abstention counted as no action, distinguishes a narrow high-performing method from a broader moderate one.

\section{Limitations and claim boundaries}\label{limitations-and-claim-boundaries}

The formal study uses one simulator and one-bucket target-field shifts; other magnitudes, interactions, persistence patterns, and non-target-field faults may yield different curves. Occupancy comes from a clean reference policy, so systems without such a stream need another maintenance design. Only 24 physical clusters are available. The exact-binomial statement is coarse and remains conditional on the exchangeable-independent component working model despite common seeds. Percentile cluster-bootstrap intervals from about 20 represented clusters may undercover in directions favorable to both structural findings. The false-admission bound is 0.20, not 0.05, and the study confirms pre-localization signal behavior, not planted-fault localization. Practical sufficiency applies only to the two frozen mask families and the 0.01-nat margin. The intended benefit is to prevent unsupported fault claims; the corresponding risk is false assurance if simulator-specific gates or thresholds are transferred to deployed or human-facing agents without new calibration.

\section{Conclusion}\label{conclusion}

Aggregate traces should be interpreted only after establishing component exposure. The two gates admitted 55/72 reference units and 54/55 matched runtime units; stable false admission was 0/20 (one-sided 95\% upper 0.1391). Affected traffic outperformed cell coverage and met the 0.01-nat sufficiency criterion within the two frozen families. Exact MHS did not improve localization over propagation-aware greedy selection; behavior under genuinely ambiguous traces remains open.

\bibliographystyle{plain}
\bibliography{references}

\end{document}